\documentclass[conference]{IEEEtran}
\IEEEoverridecommandlockouts
\usepackage{cite}
\usepackage{amsmath,amssymb,amsfonts}
\usepackage{algorithmic}
\usepackage{graphicx}
\usepackage{textcomp}
\usepackage{xcolor}
\usepackage{hyperref}
\usepackage[caption=false,font=footnotesize]{subfig}
\usepackage{multirow}
\usepackage[absolute,overlay]{textpos} 
\def\BibTeX{{\rm B\kern-.05em{\sc i\kern-.025em b}\kern-.08em
    T\kern-.1667em\lower.7ex\hbox{E}\kern-.125emX}}
    
\begin{document}

\newcommand\todo[1]{\textcolor{red}{#1}}

\title{Measuring the Ripeness of Fruit with Hyperspectral Imaging and Deep Learning\\
}

\author{\IEEEauthorblockN{Leon Amadeus Varga}
\IEEEauthorblockA{\textit{Cognitive Systems Group} \\
\textit{University of Tuebingen}\\
Tübingen, Germany \\
leon.varga@uni-tuebingen.de}
\and
\IEEEauthorblockN{Jan Makowski}
\IEEEauthorblockA{%\textit{dept. name of organization (of Aff.)} \\
\textit{LuxFlux GmbH}\\
Reutlingen, Germany \\
j.makowski@luxflux.de}
\and
\IEEEauthorblockN{Andreas Zell}
\IEEEauthorblockA{\textit{Cognitive Systems Group} \\
\textit{University of Tuebingen}\\
Tübingen, Germany \\
andreas.zell@uni-tuebingen.de}
}

\begin{textblock*}{21cm}(0.2cm, 26.7cm) % {block width} (coords)
    \noindent
    \copyright \ 2021 IEEE. Personal use of this material is permitted. Permission from IEEE must be obtained for all other uses, in any current or future media, including reprinting/republishing this material for advertising or promotional purposes, creating new collective works, for resale or redistribution to servers or lists, or reuse of any copyrighted component of this work in other works.
\end{textblock*}

\maketitle

\begin{abstract}
We present a system to measure the ripeness of fruit with a hyperspectral camera and a suitable deep neural network architecture. This architecture did outperform competitive baseline models on the prediction of the ripeness state of fruit. For this, we recorded a data set of ripening avocados and kiwis, which we make public. We also describe the process of data collection in a manner that the adaption for other fruit is easy. The trained network is validated empirically, and we investigate the trained features. Furthermore, a technique is introduced to visualize the ripening process.
\end{abstract}


\section{Introduction}
In the fruit industry, one of the goals is to determine how ripe a fruit is. Furthermore, it is helpful for supermarkets to know the ripeness level of fruit, in order not to sell far overripe fruit or give significant discounts shortly before. For fruit like bananas, the ripeness can easily be inferred from the skin color. For others like avocados, mangos, and kiwis, this is not trivial. The fruit industry mostly uses destructive indicator measurements. So only random samples are possible here. To give a solution, we verify whether hyperspectral imaging and deep neural networks could predict the ripeness level of fruit. With this work, we contribute a hyperspectral data set and tested different models on this, thereby showing the advantage of a small neuronal network.

\begin{figure}[tb]
    \centering
    \subfloat[Day 3]{\includegraphics[width=1in]{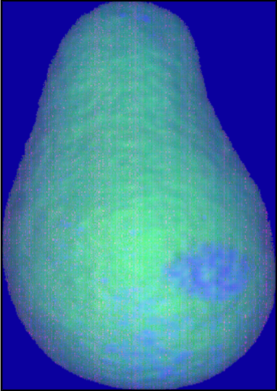}%
    \label{fig_ripening_day3}}
    \hfil
    %\subfloat[Day 4]{\includegraphics[width=1in]{images/ripening_day_4.png}%
    %\label{fig_ripening_day4}}
    %\hfil
    \subfloat[Day 5]{\includegraphics[width=1in]{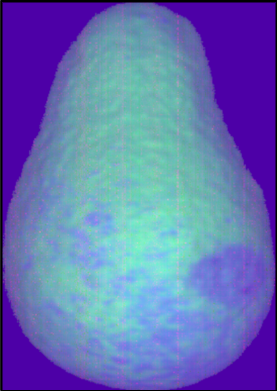}%
    \label{fig_ripening_day5}}
    \hfil
    \subfloat[Day 6]{\includegraphics[width=1in]{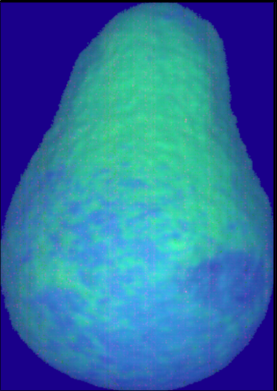}%
    \label{fig_ripening_day6}}
    \vfil
    \subfloat[Day 7]{\includegraphics[width=1in]{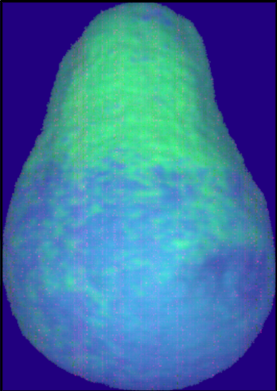}%
    \label{fig_ripening_day7}}
    \hfil
    \subfloat[Day 8]{\includegraphics[width=1in]{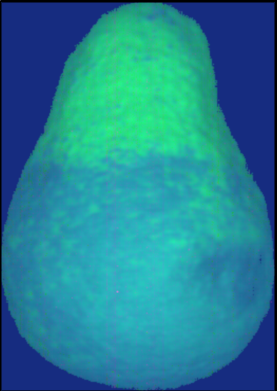}%
    \label{fig_ripening_day8}}
    \hfil
    \subfloat[Day 9]{\includegraphics[width=1in]{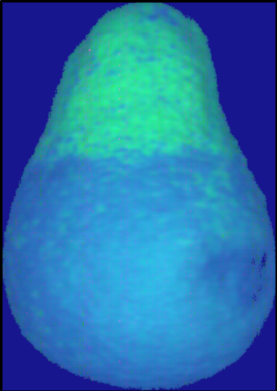}%
    \label{fig_ripening_day9}}
    \hfil
    %\subfloat[Day 10]{\includegraphics[width=1in]{images/ripening_day_10.png}%
    %\label{fig_ripening_day10}}
    %\hfil
    \caption{Visualization of the ripening process of an avocado}
    \label{fig_ripening}
\end{figure}

\section{Background and related work}
This work covers the idea of determining the ripeness level of fruit by using hyperspectral recordings. Other works already showed that it is possible to predict the ripeness of fruit by this kind of data. Pinto et al. \cite{Pinto2019} and Olarewaju et al. \cite{Olarewaju2016Non-destructiveModels} used hyperspectral imaging to determine the ripeness level of avocados. Zhu et al. \cite{Zhu2017HyperspectralModels} predicted the firmness and the soluble solids content of kiwis with hyperspectral recordings. In these three works, the authors use approaches without neural networks. So far, most fruit classification data analysis has been done with classical machine learning algorithms, which were often supported by small data sets.
In contrast to these works, we concentrate on deep learning approaches.\\
The combination of hyperspectral data and deep learning was already heavily examined in the area of remote sensing. Chen et al. \cite{Chen2014DeepData} introduced deep learning into hyperspectral remote sensing. In \cite{Makantasis2015DeepNetworks}, a convolutional neural network outperformed SVM approaches in the classification of hyperspectral remote sensing data. Ma et al. \cite{Ma2015HyperspectralLearning} used contextual deep learning for feature mining.  In \cite{Zhang2020HTD-Net:Imagery}, the HTD-Net framework was presented, which focuses on target detection with hyperspectral data. Here an autoencoder enhances the training data to produce more reliable predictions.\\
However, the use-cases of remote sensing differ widely from the classification task of fruit, and a direct comparison is not possible. \\
Mollazade et al. \cite{Mollazade2012SpatialNetworks} showed the prediction capability of a simple neural network for the moisture content of tomatoes. Gao et al. \cite{Gao2020} could predict the ripeness state of strawberries with hyperspectral imaging and a pretrained AlexNet, which is a deep convolutional neural network \cite{Krizhevsky2017ImageNetNetworks}. The ideas of both works are very similar to ours. In contrast to them, we focus on two new fruits, avocados and kiwis. For both it was already validated, that a prediction with hyperspectral data is possible \cite{Pinto2019}\cite{Olarewaju2016Non-destructiveModels}\cite{Zhu2017HyperspectralModels}.In contrast to the mentioned works we used a larger variety of models and recorded a large data set, which we make public. We further analyzed if hyperspectral data is necessary for this task or if pure color images are sufficient. The other works missed this validation.

\subsection{Hyperspectral Imaging (HSI)}
Hyperspectral imaging is a non-destructive measurement technique that has become increasingly popular in recent years. 
It is based on camera recordings that have a spectrum beyond the visible light. In contrast to standard RGB color images with three color channels, a hyperspectral image has significantly more channels \cite {sun2010hyperspectral}, usually more than 100. Each channel represents the intensity for a specific wavelength. Hyperspectral imaging can be seen as a spatial spectroscopy. 
The wavelengths around the visible range are divided into subcategories. The relevant ranges for the most frequently used hyperspectral cameras are named in Table \ref{tab:wavelengths}.
The ranges of the wavelengths reveal different chemical properties of the inspected substance. For example, the NIR range shows the presence of hydroxyl groups \cite{Mitsui2008MonitoringSpectroscopy}. Hydroxyl groups are an essential part of organic chemistry. With this in mind, it is obvious why the NIR range is vital for fruit inspection.
The application of hyperspectral imaging is very variable. Besides the named fruit inspection and remote sensing, for example, medical technology \cite{Lu2014MedicalReview} or the recycling industry \cite{Serranti2015HyperspectralControl} uses hyperspectral measurements. All applications have in common that wavelengths outside the visible range can give valuable information.\\ 
In this work, hyperspectral imaging is used to predict the ripeness level of avocados and kiwis in a non-destructive way.

\begin{table}[!t]
    \renewcommand{\arraystretch}{1.3}
    \caption{Typical wavelength ranges for hyperspectral cameras}
    \label{tab:wavelengths}
    \centering
    \begin{tabular}{l|l|l|l}
Name       & \begin{tabular}[c]{@{}l@{}}Ultra-\\ violet (UV)\end{tabular} & Visible (VIS) & \begin{tabular}[c]{@{}l@{}}Near-\\ infrared (NIR)\end{tabular} \\ \hline
Wavelength & 100-380 nm                                                   & 380-740 nm    & 740-2500 nm                                                   
\end{tabular}
\end{table}

\subsection{Fruit ripening}
Here we give a short overview of the ripening process of fruit. There is a distinction between non-climacteric and climacteric fruit. Non-climacteric fruit do not ripen after harvesting \cite{Alexander}. Therefore, the focus in the following is on climacteric fruit. 
The chemical ripening process highly depends on the fruit type.  The three main processes are \cite{Toivonen2008BiochemicalVegetables}:
\begin{itemize}
    \item Deconstruction of the cell walls, so the fruit becomes softer.
    \item Starch hydrolyzes to sugar, which leads to sweetness.
    \item Deconstruction of chlorophyll and synthesis of other pigments leads to a color change.
\end{itemize}
With the following indicators, the ripeness level of fruit are commonly measured:
\begin{itemize}
    \item Soluble Solids Content (SSC) is based on the creation of sugar while ripening. Sugars are the majority of soluble solids in most fruit.
    \item Fruit flesh firmness shows the degeneration of the cell walls. 
    \item Starch content indicates the degeneration of the starch.
\end{itemize}
Especially SSC and the fruit flesh firmness are widely used because they are reliable indicators  for many fruit types. Nowadays, their measurement is destructive. So, it is only possible to measure random samples.
Other works already showed that it is possible to predict the ripeness level by using hyperspectral imaging for some fruit \cite{Pinto2019}\cite{Olarewaju2016Non-destructiveModels}\cite{Zhu2017HyperspectralModels}.
In this work, the focus lies on two types of fruit. Avocados and kiwis are both fruit with a critical ripening process. The time window between unripe and overripe is small for both. Accordingly, the focus of this work lies at the end of the ripening process. Our goal is to predict the perfect consumption date.

\subsubsection{Avocado}
The avocado is the berry of an evergreen laurel plant. There are more than 400 different types of avocado, Hass and Fuerte being the most common. Through this broad diversity of species, the appearance of avocados may vary widely. Avocados only ripe after harvesting because the tree produces an inhibitor that prevents the fruit from ripening \cite{Lewis1978TheReview}. Besides the small consumption window, the avocado was chosen because of its relatively high price. Pinto et al. and Olarewaju et al. showed that it is possible to conclude the ripeness level with hyperspectral imaging \cite{Pinto2019}\cite{Olarewaju2016Non-destructiveModels}. Nevertheless, the most common ripeness measurement technique for avocados is the firmness of the fruit flesh.

\subsubsection{Kiwi}
Like the avocado, the looping berry fruit plant kiwi has many subspecies. The best known are probably Actinidia deliciosa and Actinidia chinensis. Their appearance is very similar, only the color of the fruit flesh differs. Hyperspectral imaging for the ripeness determination of kiwis is uncommon. Useful indicators for the ripeness of kiwis are the SSC and the firmness of the fruit flesh \cite{Martinsen1998MeasuringSpectroscopy}.

\begin{figure}[tb]
    \centering
    \subfloat[Avocados]{\includegraphics[width=1.5in]{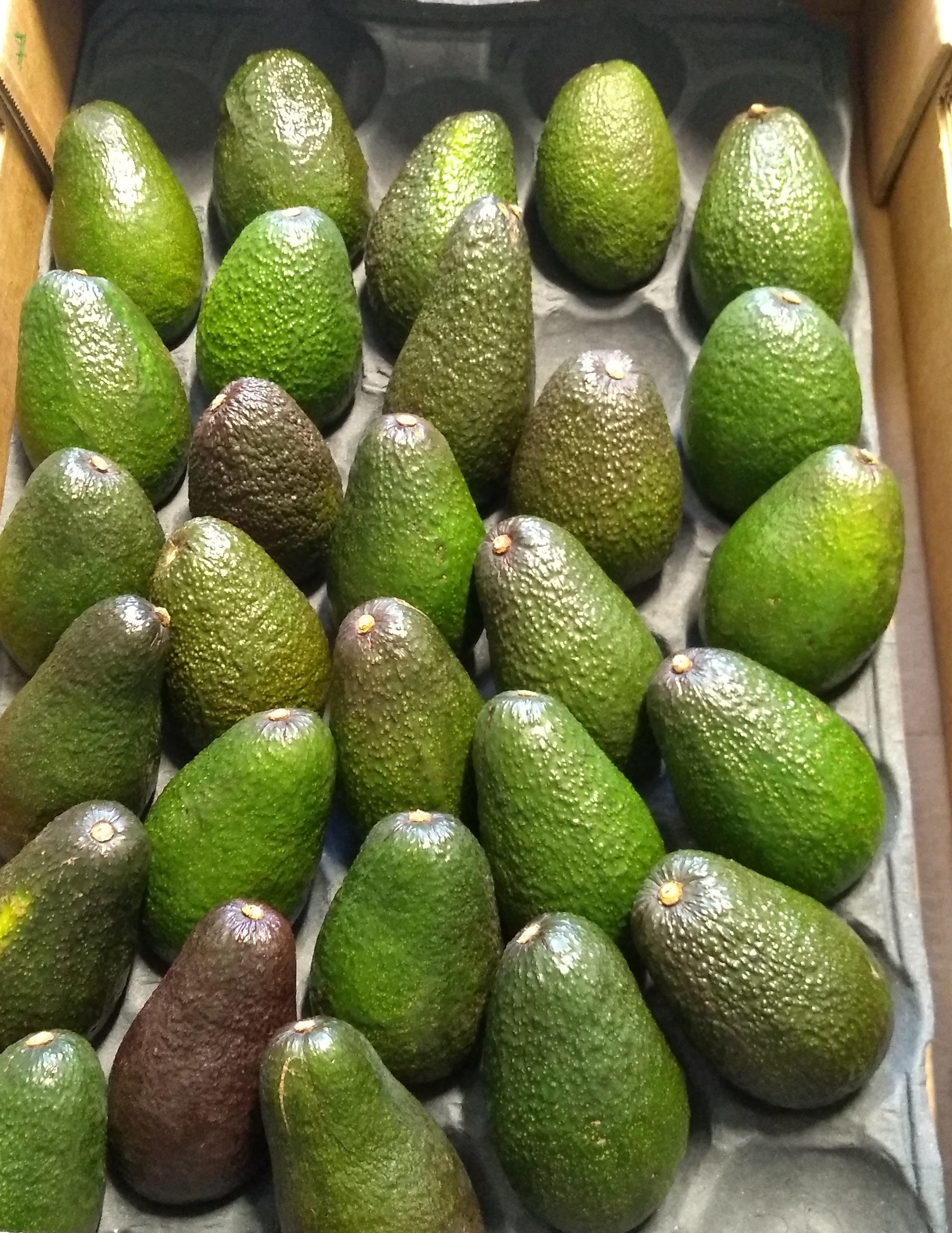}%
    \label{fig_avocados}}
    \hfil
    \subfloat[Kiwis]{\includegraphics[width=1.5in]{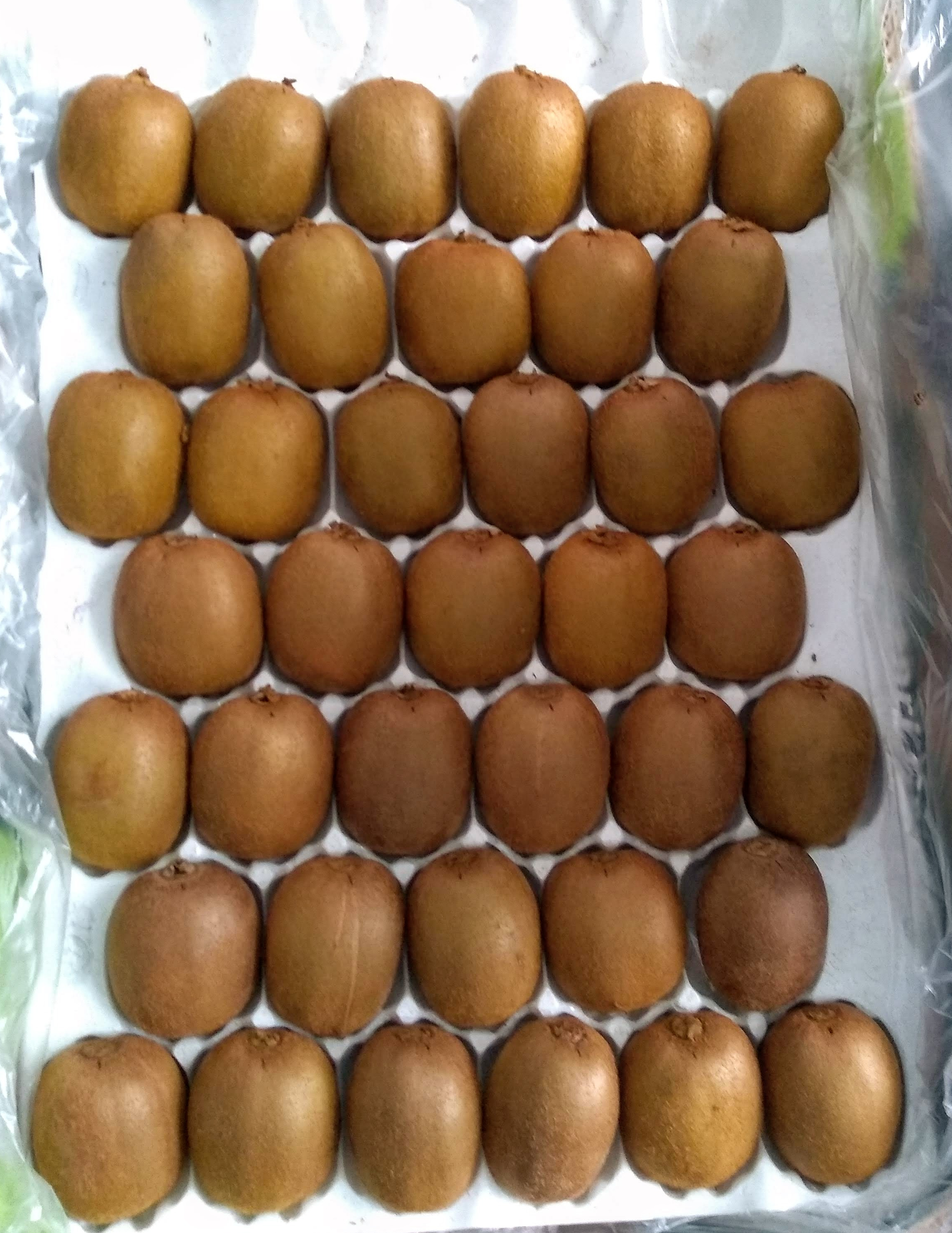}%
    \label{fig_kiwis}}
    \caption{Two of the fruit crates at day 1 of the first test series.}
    \label{fig_fruits}
\end{figure}

\section{Data set}
We now describe the measurement setup, so it is possible to reproduce the data. It is also possible to adapt the procedure for other fruit. Our hyperspectral recordings are available under \href{https://github.com/cogsys-tuebingen/deephs\_fruit}{https://github.com/cogsys-tuebingen/deephs\_fruit}. This data set is used in the further analysis. The data set contains 1038 recordings of avocados and 1522 recordings of kiwis. It covers the ripening process from unripe to overripe for both fruit. Because of the destructive manner of the labeling process, only 180 avocado recordings and 262 kiwis recordings are labeled by indicator measurements. The data set was recorded in two separate measurement series. We applied a division into training set ($\frac{3}{4}$), validation set ($\frac{1}{8}$) and test set ($\frac{1}{8}$), evenly distributed among the different states of ripening.

\begin{figure}[tb]
\centering
\includegraphics[width=\linewidth]{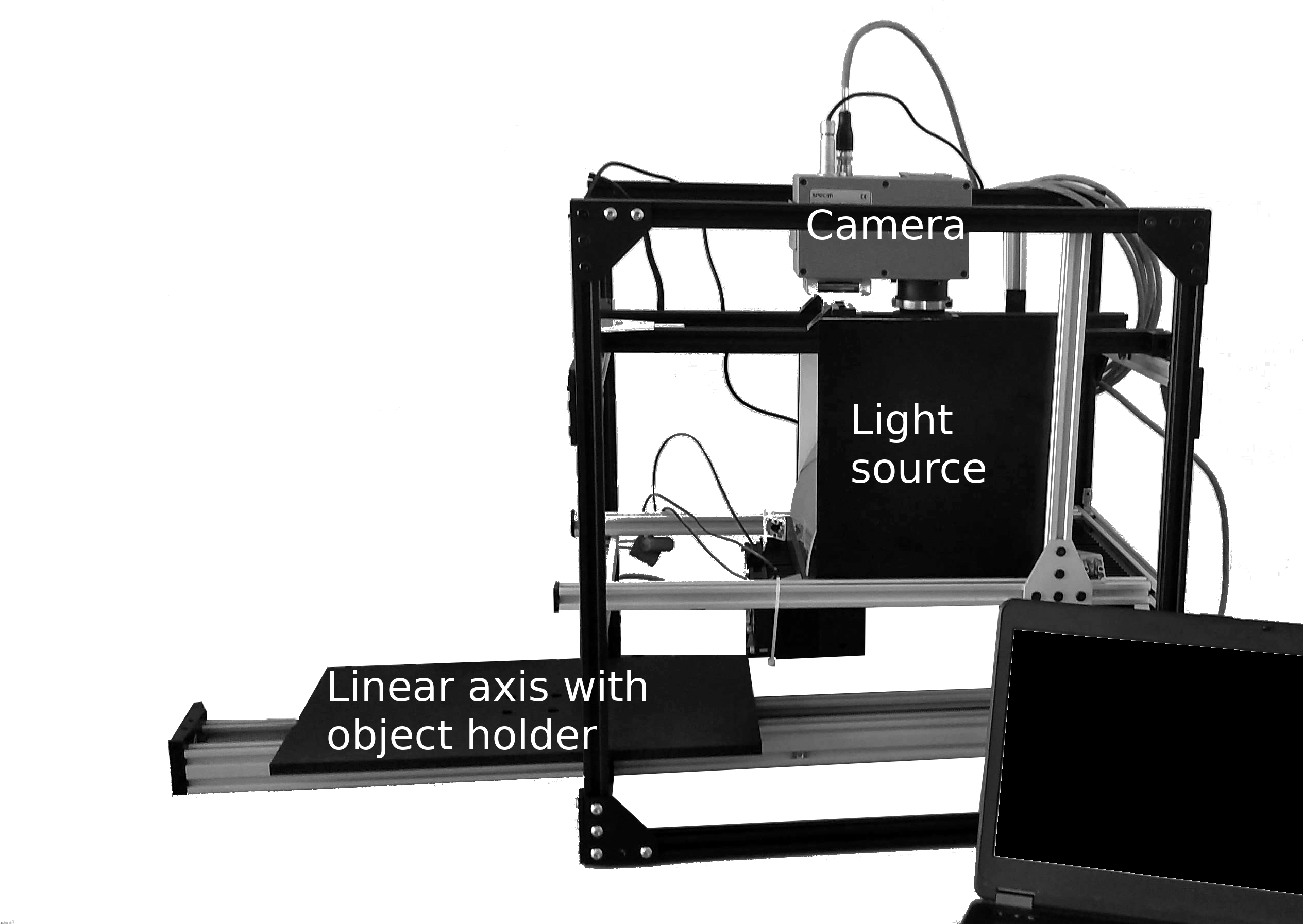}
    \caption{The recording system. With the object holder and linear axis, the light source and the camera.}
\label{fig:system}
\end{figure}

\subsection{Measurement setup}
There are three main components, which are visible in Figure \ref{fig:system}. The first component is the object holder, which is moved by a linear actuator. The linear actuator is necessary for the line scan operation mode of the hyperspectral cameras. The linear axis would not be needed for one-shot or mosaic hyperspectral cameras instead of line scan cameras. The latter, however, still seem to have better sensitivity.\\
The second component is the light source. For hyperspectral imaging, a sufficiently bright and homogeneous light source is indispensable. We used halogen lamps and LED lamps in combination to cover a broad spectrum. In addition, we used a polytetrafluoroethylene curvature reflector to create diffuse light, which is preferable.\\
The last component is the camera. We used two different cameras to allow a better validation of the results and cover various wavelength ranges. We used a Specim FX 10 and an INNO-SPEC Redeye 1.7. Both cameras operate in the line scan mode. For the second measurement series, only the Specim FX 10 was used.
The Specim FX 10 has 224 channels, and a spectral range from 400 to 1000 nm. This range holds the VIS range with the addition of the lower NIR range.
The INNO-SPEC Redeye 1.7 records 252 channels. Their spectral range run from 950 to 1700 nm.\\
Aside from that, we used a refractometer to measure the soluble solids content. A refractometer can indicate the concentration of a certain substance in the sample. For the fruit flesh firmness, we used a penetrometer. A penetrometer can measure penetration resistance. Both techniques are destructive, so only random samples as labels were possible.

\subsection{Data acquisition}
The described setup was used for two measurement series covering a total of 28 days in the years 2019 and 2020. The design of the two measuring systems used was identical and followed the described setup. We acquired fresh avocados and kiwis for the two series from a supermarket which was supporting our measurement plans. Each day the following procedure was followed:
\begin{enumerate}
    \item Record the temperature
    \item Start the measurement setup for the warm up of the lamps
    \item Calibrate the linear actuator
    \item For both cameras:
    \begin{enumerate}
        \item Adjust the focus of the camera on the surface of a reference object
        \item Record a white reference (average of 10 measurements)
        \item Record a dark reference (average of 10 measurements)
        \item Record the front and the back of each fruit to double the data without much effort
    \end{enumerate}        
    \item Select fruit for destructive indicator measurement
    \begin{enumerate}
        \item Weigh the fruit
        \item Determine the fruit flesh firmness with a penetrometer
        \item (Only for kiwis:) Measure the sugar content via the 
refractometer
        \item Record the overall ripeness level of the fruit by appearance and taste
    \end{enumerate}
\end{enumerate}
The number of destructively measured fruit was adapted to the ripening progress each day.
The output of the test series is a collection of hyperspectral recordings of kiwis and avocados. Each recording contains only one fruit.

\begin{figure*}[tb]
\centering
\includegraphics[width=0.9\textwidth]{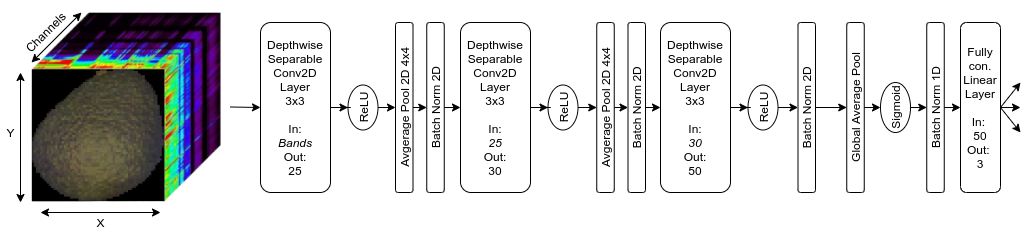}
    \caption{Architecture of our hyperspectral convolutional neural network. The image of the input cube is a adapted version of \cite{Arbeck2013}.}
\label{fig:cnn_architecture}
\end{figure*}

\subsection{Data preparation}
To improve the quality of the recorded data, we used background extraction. We excluded the background with a simple pixel-based neural network that we trained to differentiate between background and fruit. Further, the smallest possible rectangle around the fruit was extracted from the recordings to remove most of the background. We observed that the results are better if the intensity of the remaining background is forced to zero. Therefore, the results are the smallest possible recordings of the fruit with an empty background.\\
For the labels, we defined categories. Our goal was to classify whether the fruit is unripe, ripe, or overripe. Consequently, a regression problem is not necessary, and we reduced the complexity to three classes for the firmness, the sweetness, and the overall ripeness level. For the category firmness, the classes were based on the penetrometer measurements. The sweetness category, which is only useful for kiwis, was based on the refractometer tests. The last category, the ripeness, was based on the appearance and the taste. The class assignments are visible in Table \ref{tab:category_classes_avocado} and \ref{tab:category_classes_kiwi}.

\begin{table}[tb]
    \renewcommand{\arraystretch}{1.3}
    \caption{Classes for Avocados}
    \label{tab:category_classes_avocado}
    \centering
    \begin{tabular}{l|l|l|l}
Firmness  & \begin{tabular}[c]{@{}l@{}}Too hard\\ \textgreater 1200 $\frac{g}{cm^2}$\end{tabular} & Perfect & \begin{tabular}[c]{@{}l@{}}Too soft\\ \textless 900 $\frac{g}{cm^2}$\end{tabular} \\ \hline
Ripeness  & Unripe & Perfect & Overripe 
\end{tabular}
\end{table}

\begin{table}[tb]
    \renewcommand{\arraystretch}{1.3}
    \caption{Classes for Kiwis}
    \label{tab:category_classes_kiwi}
    \centering
    \begin{tabular}{l|l|l|l}
Firmness  & \begin{tabular}[c]{@{}l@{}}Too hard\\ \textgreater 1500 $\frac{g}{cm^2}$\end{tabular} & Perfect & \begin{tabular}[c]{@{}l@{}}Too soft\\ \textless 1000 $\frac{g}{cm^2}$\end{tabular} \\ \hline
Sweetness & \begin{tabular}[c]{@{}l@{}}Not sweet\\ \textless 15.5 $^\circ$Brix\end{tabular}                      & Perfect & \begin{tabular}[c]{@{}l@{}}Too sweet\\ \textgreater 17 $^\circ$Brix\end{tabular}                  \\ \hline
Ripeness  & Unripe & Perfect & Overripe 
\end{tabular}
\end{table}

\section{Experiment}
In this section we compare different models on our data set. First, we describe a simple neural network, which was designed for this application. The focus was here to reduce the chance of overfitting, so a tiny convolutional neural network was the goal. Afterwards, the training and test process is specified. Then we compare the different models. Our implementations can be found at \href{https://github.com/cogsys-tuebingen/deephs\_fruit}{https://github.com/cogsys-tuebingen/deephs\_fruit}.

\subsection{Our Hyperspectral Convolutional Neural Network}
Our Hyperspectral Convolutional Neural Network is a small neural network, which was specialized for the application of ripening fruits. We try to give reasons for some architecture decisions and explain why their usage is beneficial for hyperspectral data.\\
An RGB color image is a cuboid with two spatial dimensions and one channel dimension with the three channels red, green, blue. A hyperspectral image has significantly more channels than a color image, so the input data is much larger for the same spatial resolution. For computational reasons, it is essential to extract the necessary information at an early step. In many approaches this is done by a preprocessing step, where the most important bands are extracted and used for the further inspection. We wanted to give the network the option to select the most informative bands on its own.\\
Besides the large data size of individual images, a further problem of hyperspectral data is often a small data set in comparison to common image classification, which can lead to overfitting. Unlike for standard color images, there are no commonly usable large data sets available to train hyperspectral models.\\
In Figure \ref{fig:cnn_architecture}, the architecture of our HS-CNN for fruit classification is presented. The whole network is designed to keep it as simple and small as possible. The input is a hyperspectral recording of a fruit. The recording consists of two spatial dimensions and the channel dimension. Three convolutional layers extract feature maps from the input. The convolutions are separated into two smaller separable convolutions to reduce the number of parameters \cite{Guo2019}. Instead of the frequently used max-pooling layer, we used average-pooling layers because they gave empirically better results in our experiments. An explanation might be that for this task, the winner-takes-all strategy of max-pooling layers is counter-productive. Furthermore, batch normalization was used to speed up the training process \cite{Ioffe2015BatchShift}.
The final classification happens in the head of the CNN, consisting of a global average pooling layer and a fully connected layer. The global average pooling layer reduces the number of parameters massively and leads to more stable predictions compared to a fully connected head of similar size \cite{Lin2014}. In the present case, the network classified three different categories visible in the output of the final layer. We developed this architecture for hyperspectral recordings with around 200 channels of wavelengths. If the number of channels differs, adaptions of the hidden layers are necessary.

\begin{table*}[tb]
    \caption{Test accuracy over all categories}
    \label{tab:test_accuracy}
    \centering
   \begin{tabular}{ll|l|l|l|l|ll|l|l|l|l|}
Fruit &  & \multicolumn{4}{l|}{Avocado} & \multicolumn{6}{l|}{Kiwi} \\ \hline
Category &  & \multicolumn{2}{l|}{Firmness} & \multicolumn{2}{l|}{Ripeness} & \multicolumn{2}{l}{Firmness} & \multicolumn{2}{l|}{Sweetness} & \multicolumn{2}{l|}{Ripeness} \\
Camera &  & \begin{tabular}[c]{@{}l@{}}INNO-\\ SPEC\\ Redeye\end{tabular} & \begin{tabular}[c]{@{}l@{}}Specim\\ FX 10\end{tabular} & \begin{tabular}[c]{@{}l@{}}INNO-\\ SPEC\\ Redeye\end{tabular} & \begin{tabular}[c]{@{}l@{}}Specim\\ FX 10\end{tabular} & \multicolumn{1}{l|}{\begin{tabular}[c]{@{}l@{}}INNO-\\ SPEC\\ Redeye\end{tabular}} & \begin{tabular}[c]{@{}l@{}}Specim\\ FX 10\end{tabular} & \begin{tabular}[c]{@{}l@{}}INNO-\\ SPEC\\ Redeye\end{tabular} & \begin{tabular}[c]{@{}l@{}}Specim\\ FX 10\end{tabular} & \begin{tabular}[c]{@{}l@{}}INNO-\\ SPEC\\ Redeye\end{tabular} & \begin{tabular}[c]{@{}l@{}}Specim\\ FX 10\end{tabular} \\
 &  &  &  &  &  & \multicolumn{1}{l|}{} &  &  &  &  &  \\ \hline
\textbf{SVM} &  & 77.8\% & 73.3\% & 44.4\% & 66.7\% & \multicolumn{1}{l|}{44.4\%} & 60.9\% & 44.4\% & \textbf{82.6\%} & 33.3\% & 45.8\% \\ \hline
\textbf{kNN} &  & 73.3\% & 77.8\% & \textbf{88.9\%} & 60.0\% & \multicolumn{1}{l|}{\textbf{55.6\%}} & 60.9\% & 22.2\% & 73.9\% & 55.6\% & 50.0\% \\ \hline
\multirow{3}{*}{\begin{tabular}[c]{@{}l@{}}\textbf{ResNet-18}\\ \\ 11M parameters\end{tabular}} & RGB & 66.7\% & 66.7\% & 66.7\% & 53.3\% & \multicolumn{1}{l|}{44.4\%} & 56.5\% & 55.6\% & 47.8\% & 44.4\% & 54.2\% \\ \cline{2-12} 
 & PCA & 44.4\% & 53.3\% & 44.4\% & 60.0\% & \multicolumn{1}{l|}{33.3\%} & 60.9\% & 44.4\% & 47.8\% & 66.7\% & 33.3\% \\ \cline{2-12} 
 & Full & 66.7\% & 80.0\% & 33.3\% & 80.0\% & \multicolumn{1}{l|}{\textbf{55.6\%}} & 60.9\% & \textbf{66.7\%} & 47.8\% & 66.7\% & 58.3\% \\ \hline
\multirow{3}{*}{\begin{tabular}[c]{@{}l@{}}\textbf{AlexNet}\\ \\ 58M parameters\end{tabular}} & RGB & 44.4\% & 33.3\% & 33.3\% & 33.3\% & \multicolumn{1}{l|}{33.3\%} & 52.2\% & 44.4\% & 47.8\% & 33.3\% & 33.3\% \\ \cline{2-12} 
 & PCA & 44.4\% & 33.3\% & 33.3\% & 33.3\% & \multicolumn{1}{l|}{33.3\%} & 52.2\% & 44.4\% & 47.8\% & 33.3\% & 33.3\% \\ \cline{2-12} 
 & Full & 44.4\% & 33.3\% & 33.3\% & 60.0\% & \multicolumn{1}{l|}{33.3\%} & 52.2\% & 44.4\% & 47.8\% & 66.7\% & 33.3\% \\ \hline
\multirow{3}{*}{\begin{tabular}[c]{@{}l@{}}\textbf{HS-CNN (our)}\\ \\ 32K parameters\end{tabular}} & RGB & 77.8\% & 53.3\% & 55.6\% & 40.0\% & \multicolumn{1}{l|}{44.4\%} & 65.2\% & 55.6\% & 60.9\% & 44.4\% & 62.5\% \\ \cline{2-12} 
 & PCA & 44.4\% & 80.0\% & 44.4\% & 66.7\% & \multicolumn{1}{l|}{44.4\%} & 34.78\% & 44.4\% & 47.8\% & 33.3\% & 33.3\% \\ \cline{2-12} 
 & Full & \textbf{88.9\%} & \textbf{93.3\%} & \textbf{88.9\%} & \textbf{93.3\%} & \multicolumn{1}{l|}{44.4\%} & \textbf{69.57\%} & \textbf{66.7\%} & \textbf{82.6\%} & \textbf{77.8\%} & \textbf{66.7\%} \\ \hline
\end{tabular}
\end{table*}

\subsection{Training}
For training, the size of the classes in the categories was balanced. Thus, there was no bias towards one class. We used rotation, flipping, random noise and random cut as data augmentation techniques, as each of these doesn't change the label. The neural networks were optimized with Adabound using $1 \times 10^{-2}$ as learning rate \cite{Luo2019AdaptiveRate}. Focal loss was used as loss function \cite{Lin2017FocalDetection}. We used early stopping based on the validation loss to prevent over-fitting \cite{Prechelt1998AutomaticCriteria}. For training, we used a batch size of 32. The hyperspectral images were resized to 64x64 pixels.

\subsection{Test}
We tested five models on our data set. The models were a Support Vector Machine (SVM) with a radial basis kernel \cite{Cristianini2000AnMethods}, a k-nearest neighbor classifier (kNN) \cite{Fix1989DiscriminatoryProperties} and a ResNet-18, a convolutional neural network architecture with identity shortcut connections and 18 layers \cite{He2016DeepRecognition}. Further an AlexNet, which performed well for strawberries \cite{Gao2020}, and our Hyperspectral Convolutional Neural Network (HS-CNN) were used. The ResNet-18 was used, because a larger represenantive of the ResNet-family would more likely tend to over-fitting. For the ResNet-18 and the AlexNet, the first layer of the network was adapted to the hyperspectral images as input. The parameter \textit{C} of the SVM was evaluated by grid search with cross validation on the training set. The same applies to the parameter \textit{k} of the kNN.\\
The test-set was $\frac{1}{8}$ of the labeled hyperspectral recordings. For the evaluation test time augmentation \cite{Howard2014SomeClassification} was used. The test results are given in Table \ref{tab:test_accuracy}. For each neural network three values are given. The \emph{Full} value gives the accuracy when the network has access to the whole hyperspectral recording. In the \emph{RGB} case, the hyperspectral recordings were reduced to color images in a preprocessing step. And for the \emph{PCA} case, a Principal Component Analysis (PCA) was used to reduce the channel size of the hyperspectral recordings to five. The PCA technique is often used for hyperspectral recordings to extract only the necessary information in an early step.\\
Our model outperformed the reference models in most cases. Moreover, it produced the most stable results. With our model, it was possible to predict the firmness of avocados with an accuracy of over 93.33 \% and further predict the ripeness level in 3 categories with over 90 \%. The prediction of the ripeness level of the kiwis is much harder than for the avocados. Thus, the prediction accuracy for them was significantly lower for all models. However, our model could still predict the firmness of untested kiwis with an accuracy of nearly 70\% and the ripeness with nearly 80\%. Further the \emph{Full} use case was in most cases better than the reduced use cases (\emph{RGB} or \emph{PCA}). In the \emph{Full} case the network could select the most influential bands. \emph{RGB} was in some cases better than the \emph{PCA} approach. The \emph{RGB} reduction doesn't use the largest variance in contrast to \emph{PCA}. Instead, it uses the CIE color matching functions to calculate the impact of each wavelength. Most likely \emph{PCA} removes some necessary information by the reduction, which are still available in the \emph{RGB} reduction.

%\begin{figure}[tb]
%    \centering
%    \includegraphics[width=3in]{images/confusion_matrix_avocado.png}
    %\subfloat[Kiwis NIR]{\includegraphics[width=3in]{images/confusion_matrix_kiwi.png}%
    %\label{fig_confusion_matrix_kiwi}}
%    \caption{Confusion matrix for avocado ripeness (Specim FX 10)}
%    \label{fig_confusion_matrix}
%\end{figure}

\begin{figure}[tb]
    \centering
    \subfloat[Day 3]{\includegraphics[width=1.5in]{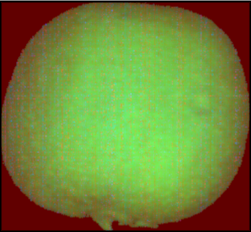}%
    \label{fig_firmness_day3}}
    \hfil
    \subfloat[Day 4]{\includegraphics[width=1.5in]{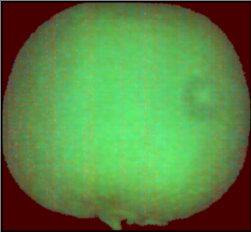}%
    \label{fig_firmness_day4}}
    \vfil
    \subfloat[Day 5]{\includegraphics[width=1.5in]{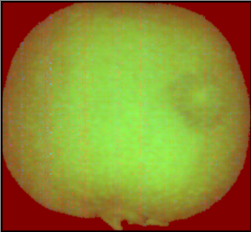}%
    \label{fig_firmness_day5}}
    \hfil
    \subfloat[Day 6]{\includegraphics[width=1.5in]{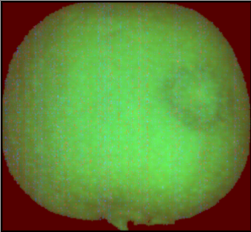}%
    \label{fig_firmness_day6}}
    \caption{Visualization of the firmness distribution of a kiwi}
    \label{fig_firmness}
\end{figure}

\begin{figure*}[tb]
    \centering
    \subfloat[Spatial based]{\includegraphics[width=3in]{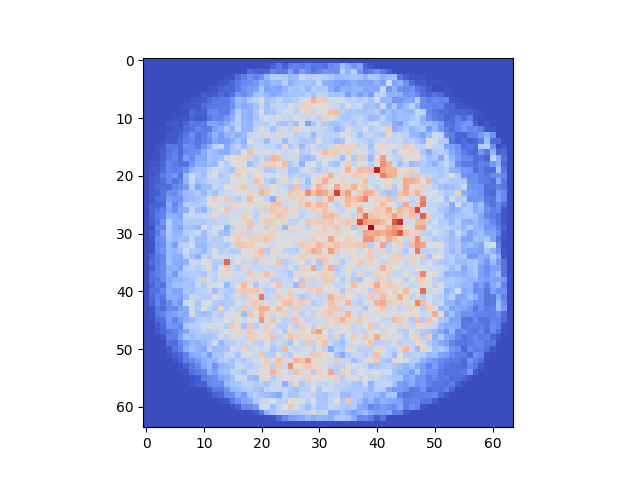}%
    \label{fig_impact_pixel}}
    \hfil
    \subfloat[Wavelength based]{\includegraphics[width=3in]{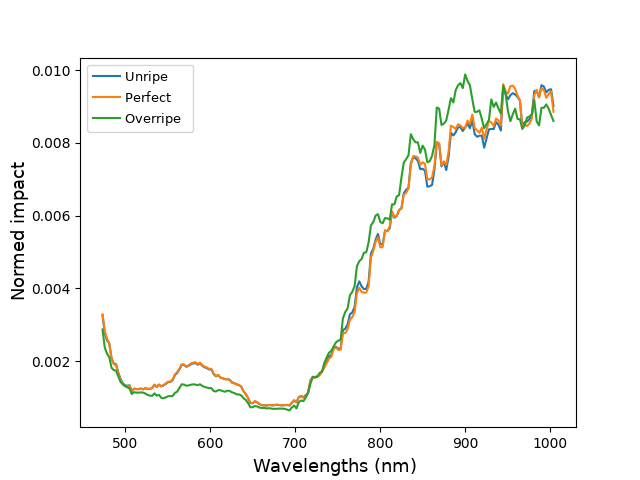}%
    \label{fig_impact_wavelength}}
    \caption{The impact of the input on the decision of the class for an avocado recorded with the Specim FX 10.}
    \label{fig_impact}
\end{figure*}

\section{Ablation study}
By removing or replacing components of our Hyperspectral Convolutional Neural Network, we study the impact of the different parts. In the following, the test accuracy for the prediction of the avocado firmness is given.

\subsection{Augmentation}
The influence of the different augmentation techniques is visible in this table. Random cut and test time augmentation seems essential in this scenario. On the other hand, the effect of the  transformation augmentations is smaller, so fruit alignment seems to be less of an issue in this data set.\\

\begin{tabular}{l|l}
\centering
Augmentation variant & Accuracy \\ \hline
\emph{Full augmentation} & \emph{93.3 \%}\\ \hline
Without test time augmentation & 70.8 \% \\ \hline
Without random noise & 73.3 \% \\ \hline
Without random cut & 69.3 \% \\ \hline
%No augmentation & 76.00 \% \\ \hline
\begin{tabular}[c]{@{}l@{}}No transformation \\ augmentation\end{tabular} & 80.0 \% 
\end{tabular}

\subsection{Depth-wise separable convolution (DSCNV)}
The idea behind depth-wise separable convolution \cite{Guo2019} is to split up the normal convolution into the spatial and a depth-wise convolution, which corresponds to the channel dimension. With this technique, the number of parameters is reduced, which can prevent overfitting.\\

\begin{tabular}{l|l|l}
\centering
Convolution type & Normal convolution & \emph{DSCNV} \\ \hline
Accuracy & 80.0\% & \emph{93.3 \%}
\end{tabular}

\subsection{Head}
The head of the network uses the feature map of the convolutional part to determine the classification result. We inspected three head architectures. A fully connected head, a Global Average Pooling \cite{Lin2014} head and a head based on Global Average Pooling with an additional linear layer. The Global Average Pooling reduces the number of parameters, which prevents overfitting. Still an additional linear layer is useful in this case.\\

\begin{tabular}{l|l}
Head architecture & Accuracy \\ \hline
\begin{tabular}[c]{@{}l@{}}\emph{Global Average Pooling}\\ \emph{with additional layer} \cite{Lin2014}\end{tabular} & \emph{93.3 \%} \\ \hline
Global Average Pooling \cite{Lin2014} & 80.0 \% \\ \hline
Fully connected layers & 86.7 \%
\end{tabular}

\subsection{Loss function}
The Focal loss is a Cross entropy loss which weights the impact of a sample corresponding to their classification error. This improves the behavior with unbalanced classes \cite{Lin2017FocalDetection}. Although we were careful to avoid class bias, the Focal loss still improves the result.\\

\begin{tabular}{l|l|l}
\centering
Loss function & Cross entropy loss & \emph{Focal loss} \\ \hline
Accuracy & 80.0\% & \emph{93.3 \%}
\end{tabular}\\

\subsection{Optimizer}
We tested different optimizers for the training process. In our case, the Adabound optimizer with a learning rate of 0.01 worked best.\\

\begin{tabular}{l|l}
\centering
Optimizer & Accuracy \\ \hline
Stochastic gradient descent \cite{Kiefer1952StochasticFunction} & 80.0 \% \\ \hline
Adam \cite{Kingma2015} & 80.0 \% \\ \hline
Adabound with default parameters \cite{Luo2019AdaptiveRate}& 80.0 \% \\ \hline
\emph{Adabound with learning rate 0.01} \cite{Luo2019AdaptiveRate}& \emph{93.3 \%}
\end{tabular}

\subsection{Pooling layers}
We compared max pooling layers with average pooling layers. The results with average pooling layers were minimally better. For this problem, it seems to be more important not to consider only the extreme value.\\

\begin{tabular}{l|l|l}
\centering
Pooling & Max pooling & \emph{Average pooling} \\ \hline
Accuracy & 86.7\% & \emph{93.3}\%
\end{tabular}

\begin{figure*}[tb]
    \centering
    \subfloat[Autoencoder]{\includegraphics[width=\linewidth]{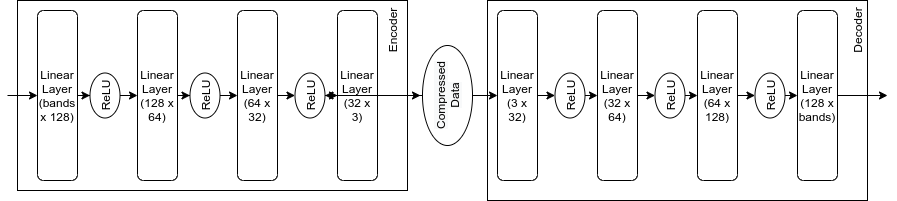}%
    \label{fig_pretrained_autoencoder}}
    \vfil
    \subfloat[Classifier network]{\includegraphics[width=\linewidth]{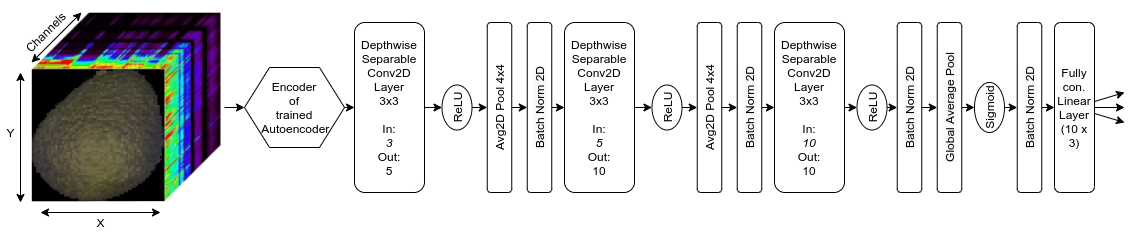}%
    \label{fig_pretrained_classifier}}
    \caption{The architecture of the Pretrained approach. The image of the input cube is a adapted version of \cite{Arbeck2013}.}
    \label{fig_pretrained}
\end{figure*}

\section{Investigation of the learned CNN features}
Besides the ablation study we want to show that the trained HS-CNN network learns meaningful features for the classification, which validates the correctness of the prediction. We used Integrated gradient\cite{Sundararajan2017AxiomaticNetworks} to see what parts of the hyperspectral recording are important to determine the state of the fruit. This technique can show the influence of neurons on the decision of the network. It is possible to validate the decision process of the neural network to a certain extent.\\
In Figure \ref{fig_impact_pixel} the spatial distribution of the impact for the avocado ripeness prediction is presented. The impact is evenly distributed over the whole fruit. In Figure \ref{fig_impact_wavelength} the wavelength-based impact is visualized. The main decision happens over 800 nm. This discovery fits with the findings of Pinto et al. \cite{Pinto2019}. Additionally, to a small extent, the range of the visible light between 520 nm and 650 nm was used by the network to differentiate between unripe and perfect fruit. This range matches the visible change of the avocados.
Overall the features learned by the convolutional neural network seem plausible.

\section{Visualization of the ripening process}
Furthermore, we introduce a technique to generate false-color images of hyperspectral recordings for specific tasks.
For this, we used a two-stage training process and a two-level classifier, presented in Figure \ref{fig_pretrained}.
In the first step, we trained a pixel-based autoencoder (Figure \ref{fig_pretrained_autoencoder}) to encode and decode hyperspectral images of fruit. The unlabeled data can also be used here. We used the mean-squared error for training. The latent space had a size of three, so the interpretation as a color image is possible. In the second step, we used the encoder's embedding as the input for a classifier network (Figure \ref{fig_pretrained_classifier}) and trained the classifier to differ between ripeness levels. Here a Focal loss was used \cite{Lin2017FocalDetection}. For the second step the labeled data is necessary. The weights of the encoder were not fixed in the second step. So, the embedding representation was adapted to fit better to the classification task. As a result, we got an encoder specialized in encoding information to differentiate ripeness levels. \\
An encoder we have trained in this way can produce false-color images that visualize the ripening process.\\
For avocados, an example is visible in Figure \ref{fig_ripening}. The ripe parts are growing from the bottom to the top of the fruit. Another example is visible in Figure \ref{fig_firmness}. Here the encoder was specialized for firmness prediction. The output visualizes the firmness distribution of a kiwi. A damaged part slowly grows over the fruit.\\
A big advantage of this technique is, that it can benefit from the large amount of unlabeled data.

\section{Conclusion}
In this work, we showed that convolutional neural networks may be used on hyperspectral data to classify exotic fruit into three classes (unripe, ripe, and overripe). We published a data set of ripening avocados and kiwis. Our HS-CNN classifier network shows superb performance in the classification of ripeness states for avocados and good performance for kiwis. We could validate the results by a more in-depth look into the trained features. Moreover, we described how to record further data. Besides that, we presented a technique to produce false-color images for specific use-cases with a pretrained autoencoder. \\
Semi-supervised approaches are particularly promising for further research, as they can also use unlabeled data sets.

\section*{Acknowledgment}
The authors would like to thank the LuxFlux GmbH company for their support with hardware and domain knowledge. %LuxFlux GmbH provided the measurement system called Polyscanner.

\bibliography{IEEEabrv,references}
\end{document}